\documentclass{article}

\PassOptionsToPackage{numbers, compress}{natbib}

\usepackage[preprint]{neurips_2020}

\usepackage[utf8]{inputenc} 
\usepackage[T1]{fontenc}    
\usepackage{hyperref}       
\usepackage{url}            
\usepackage{booktabs}       
\usepackage{amsfonts}       
\usepackage{nicefrac}       
\usepackage{microtype}      
\usepackage{graphicx}
\usepackage[noend]{algpseudocode}
\usepackage{amsmath,amssymb}
\usepackage{algorithm}
\usepackage{mathtools}
\usepackage{subcaption}

\DeclarePairedDelimiter\floor{\lfloor}{\rfloor}

\usepackage{amsthm}

\theoremstyle{definition}
\newtheorem{definition}{Definition}[section]
\newtheorem{theorem}{Theorem}[section]

\newtheorem{lemma}[theorem]{Lemma}

\newtheorem{innercustomthm}{Theorem}
\newenvironment{customthm}[1]
  {\renewcommand\theinnercustomthm{#1}\innercustomthm}
  {\endinnercustomthm}
  
\newtheorem{innercustomlem}{Lemma}
\newenvironment{customlem}[1]
  {\renewcommand\theinnercustomlem{#1}\innercustomlem}
  {\endinnercustomlem}
\usepackage{xcolor}

\newcommand{\argmin}{\arg\!\min}

\title{Tangent Space Sensitivity and Distribution of Linear Regions in ReLU Networks}

\author{%
  Balint Daroczy \\
  Institute for Computer Science and Control (SZTAKI)\\
  Kende str. 13-17, Budapest, Hungary, H-1111\\
  \texttt{daroczyb@ilab.sztaki.hu}}

\begin{document}

\maketitle

\begin{abstract}
Recent articles indicate that deep neural networks are efficient models for various learning problems. However they are often highly sensitive to various changes that cannot be detected by an independent observer. As our understanding of deep neural networks with traditional generalization bounds still remains incomplete, there are several measures which capture the behaviour of the model in case of small changes at a specific state. In this paper we consider adversarial stability in the tangent space and suggest tangent sensitivity in order to characterize stability. We focus on a particular kind of stability with respect to changes in parameters that are induced by individual examples without known labels. We derive several easily computable bounds and empirical measures for feed-forward fully connected ReLU (Rectified Linear Unit) networks and connect tangent sensitivity to the distribution of the activation regions in the input space realized by the network. Our experiments suggest that even simple bounds and measures are associated with the empirical generalization gap.
\end{abstract}

\section{Introduction}

Recent theoretical and empirical results go beyond traditional generalization bounds for deep neural networks e.g. uniform convergence \cite{sontag1998vc,bartlett2003vapnik} or algorithmic complexity \cite{liu2017algorithmic}. There are several promising ideas inspired by statistical physics \cite{rolnick2017power,lin2016does}, tensor networks \cite{stoudenmire2016supervised} or differential geometry \cite{amari1996neural,kanwal2017comparing,jacot2018neural,ay2017information}. Certain, even surprising empirical phenomenon (e.g. larger networks generalize better, different optimization with zero training error may generalize differently \cite{neyshabur2018towards}, well-performing models are accessible by short distance in the parameter space \cite{du2018gradient} or the magnitude of the initial parameters predetermine the magnitude of the learned parameters \cite{bartlett2017spectrally}) provided recent progress in theoretical explanations of generalization by core properties of the learned models. Without any claim of completeness, generalization was connected to the number and magnitude of the parameters with regularization methods \cite{krogh1992simple,neyshabur2014search,zhang2016understanding}, the depth and width of the network, initialization and optimization of the parameters \cite{srivastava2014dropout,du2018gradient,cao2019generalization}, augmentation \cite{perez2017effectiveness} or local geometrical properties of the state of the network \cite{wilson2017marginal,keskar2017improving}. Authors in \cite{liang2017fisher} suggest Fisher-Rao norm (inner product of the normalized parameter vector and the parameter vector) as a measure of generalization in case of bias-less feedforward neural networks with linear activation functions. Beside exciting theoretical advancements (e.g. flatness of the minimum can be changed arbitrarily under some meaningful conditions via exploiting symmetries \cite{dinh2017sharp}) our understanding of deep neural networks still remain incomplete \cite{neyshabur2018towards,cao2019generalization}. One of the important question is how empirical generalization gap (difference between train and test set accuracy) relates to properties of the trained models \cite{ jiang2018predicting}. 

Our motivation for examining the effect of adversarial robustness in the tangent space on feed-forward neural networks with ReLU activations is twofold: robustness of deep networks to adversarial attacks and recently discovered knowledge about ReLU networks. In this paper we consider adversarial example generation and smooth augmentation methods to check whether a model could handle small or meaningful changes of genuine examples. Due comparability, performance of newly developed models are measured on open and well-known benchmark datasets. Matter of concern, especially the best performing models suffer under adversarial attacks (missclassification in case of small changes even if the perturbations are undetectable by an independent observer) as they are highly sensitive to small changes in the data due their high complexity among others \cite{goodfellow2014explaining}. The authors in \cite{NIPS2019_9336} dispute that uniform convergence may be unable to explain generalization as the decision boundary learned by the model could be so complex it affects uniform convergence. Our motivation is based on the assumption that if the optimization handles adversarial changes in the training and the test set similarly the model may not be overfitted and the learning procedure could be less sensitive to noise or adversarial attacks. Additionally, we argue that this property can be measured at some extent without exactly measuring the loss (therefore no need for validation labels) and closely connected to the properties of the function the model realizes specifically the state of the models and the data distribution. There are several ways to address adversarial perturbations. As an example, authors in \cite{novak2018sensitivity} showed that the norm of the input-output sensitivity, the Frobenius norm of the Jacobian matrix of the output w.r.t. input has a strong connection to generalization in case of simple architectures. Recently, the authors in \cite{werpachowski2019detecting} suggested an approach to detect overfitting of the model to the test set. They use an adversarial error estimator with importance weighting (adversarial example generator (AEG)) to detect covariate shift in the data distribution while measuring independence between the model and the test set. Our hypothesis is that robustness to adversarial changes is not a uniform property thus we cannot capture in its entirety with traditional second order measures as Hessian or with first order sensitivity measures as suggested in \cite{novak2018sensitivity} thus we investigate the tangent space. 

In addition, recent results on ReLU networks suggest that the hypothesis, that deep neural networks are exponentially more efficient in regard of maximal capacity (representational power) in comparison to ``shallow'' networks, does not explain why deep networks perform better in practice as the complexity of the network measured by the number of non zero volume linear regions increases with the number of neurons independently of the topology of network \cite{hanin2019deep}. Maybe more surprisingly, under simple presumptions the number of linear regions does not increase (or decrease) throughout learning except in non-realistic cases e.g. learning with random labels or memorization. If so the question remains, what is happening during learning if the number of activation regions are not changing? An explanation consider parametrized trajectories in the input space  \cite{raghu2017expressive} and show that the trajectory lengths increase exponentially with the depth of the network measured by transitions in linear regions throughout the trajectory. Our main hypothesis is that learning may adjust the distribution of activation regions and there is a possible relation to adversarial robustness. Our contributions are the following. 


\begin{itemize}
\item We suggest a measure, \textit{tangent sensitivity} which characterizes, in a way, both the geometrical properties of the function and the original data distribution without the target meanwhile captures how the model handle injected noise at each layer per sample. In comparison to \cite{arora2018stronger} our measure operates on directional derivatives.  
\item We derive several easily computable bounds and measures for feed-forward ReLU multi-layer perceptrons based either only on the state of the network or on the data as well. Throughout of these measures we connect \textit{tangent sensitivity} to the structure of the network and particularly to the input-output paths inside the network, the norm of the parameters and the distribution of the linear regions in the input space. The bounds are closely related to path-sgd \cite{NIPS2015_5797}, the margin distribution \cite{bartlett2017spectrally,neyshabur2017exploring,jiang2018predicting} and narrowness estimation of linear regions \cite{zhang2020empirical} albeit primarily to the distribution of non zero volume activation patterns. 
\item Finally, we experiment on the CIFAR-10 \cite{krizhevsky2009learning} dataset and observe that even simple upper bounds of \textit{tangent sensitivity} are connected to empirical generalization gap, the performance difference between the training set and test set. 
\end{itemize}

The paper is organized as follows; we set notations in Section~\ref{sec:prelim}, we define \textit{tangent sensitivity} and describe our main findings in Section~\ref{sec:tss}, suggest connection to generalization in Section~\ref{sec:tss_gen} and finally, we discuss the experiments in Section~\ref{sec:results}.

\section{Preliminaries}
\label{sec:prelim}

Let $f$ be a function from the class of feed-forward fully connected neural networks with input dimension $d_{in}$, output dimension $d_{out}$ and ReLU activation functions ($\sigma(z)=\max\{0,z\}$ with $z \in \mathbb{R}$). The network structure is described as a weighted directed acyclic graph (DAG) $G(V,E)$ with $d_{in}$ input nodes $v_{in}[1],..,v_{in}[d_{in}]$, $c$ output nodes $v_{out}[1],..,v_{out}[d_{out}]$, a finite set of hidden nodes and weight parameters assigned to every edge. The network is organized in ordered layers, the input layer (elements of $v_{in}$), a set of hidden layers (disjoint subsets of hidden nodes) and the output layer (elements of $v_{out}$) without edges inside the layers. There are only out edges from a layer to the next layer thus every directed path in $G$ connecting an input and an output node has length equal to the number of layers. These paths are the longest directed paths in $G$. We refer the set of directed paths between an input and an output node in a network with depth $k$ as a set of an input-output paths: $P_{i,j}(x;\theta)=[\{v_{in}[i] \xrightarrow{w_{p_{i,j}}[1]}  h_{p_{i,j}}[1](x;\theta) \xrightarrow{w_{p_{i,j}}[2]},...,h_{p_{i,j}}[k-1](x;\theta)\xrightarrow{w_{p_{i,j}}[k]}  v_{out}[j]\}]$. If given input and the state of the network every preactivation and every weight along a path are non zero, we will call the path as an \textit{active path}. Altogether the network with depth $k$ is defined as a parametric function $f(x;\theta=\{W_i,b_i;i=\{1,..,k\}\}) = W_k[...[W_2[W_1^Tx+b_1]_++b_2]_+...]_++b_k$ where $x\in\mathbb{R}^d$ and $\forall i$ $W_i\in \mathbb{R}^{N_{i-1} \times N_{i}}, b_i \in \mathbb{R}^{N_i}$ with $N_{\theta}=|\theta|$ as the number of trainable parameters, $N_i$ as the number of hidden units in the $i$-th layer and the number of neurons as $N=\sum_i^k N_i$. We will refer the preactivation of the $l$-th neuron (the $j$-th neuron in the $i$-th layer) as $h_l(x;\theta) = h_{i,j}(x;\theta)  = [W_i[...[W_2[W_1^Tx+b_1]_++b_2]_+...]_++b_{i}]_j$. 

In addition, following the definitions in \cite{hanin2019deep}, we define an \textit{activation pattern} for a network $f$ with assigning a sign the each neuron in the network, $A = \{a_{l};l=1,..,N\} \in \{-1,1\}^N$. For a particular input we will refer $A(x;\theta) = \{sign(h_{l}(x;\theta)); l=1,..,N\}$ as the activation pattern assigned to an input $x$ and $n_i(x;\theta)$ as the number of hidden units in the $i$-th layer with positive activations (their value in the activation pattern is $1$). An \textit{activation region} with the corresponding fixed $\theta$ and $A$ is defined as $R(A;\theta):=\{x \in \mathbb{R}^{d_{in}} | sign(h_{l}(x;\theta)) = a_{l}\}$, the set of input assigned to the same activation pattern. The non-empty activation regions are the \textit{activation regions} of $f$ at $\theta$. In comparison, linear regions of a network at state $\theta$ are the input regions where the function defines different linear regions. The number of activation regions are higher or equal to the number of linear regions e.g. if the transitions between two neighbouring activation region the function is continuous in $\nabla f$ thus they belong to the same linear region (for more detail see Lemma 3 in \cite{hanin2019deep}). It is worth mentioning that linear regions are not necessary convex however activation regions are convex (see Theorem 2 in \cite{raghu2017expressive}). 

We consider the problem of \textit{Empirical Risk Minimization}, where given a finite set of samples $\{(x_i,y_i) ; i=\{1,..,n\}\}$ drawn from a probability distribution $D$ on $\Omega \times \{-1,1\}$ we minimize the empirical loss, $L_{emp}(f)$ over the elements in a previously chosen function class $f \in \mathcal{F}$ as $f^* =  \argmin_{f \in \mathcal{F}} L_{emp}(f) := \argmin_{f \in \mathcal{F}} \sum_{i=1}^n l(f(x_i),y_i)$ an approximation of $\argmin_{f \in \mathcal{F}}\mathbf{E}_{x,y \sim D} [l(f(x_i),y_i)]$. We will refer to the difference between the empirical loss on the training set and on the test set as \textit{empirical generalization gap} to differentiate it from the \textit{generalization gap} where the difference is taken between the empirical loss and a true loss. They have a natural connection, for more see e.g. the proof of Vapnik-Chervonenkis theorem in Chapter 12 in \cite{devroye2013probabilistic}. Neural networks are typically trained by first or second order gradient descent methods over the parametrized space $\Theta$ with $N_{\theta}$ parameters. These iterative methods often produce local minimums as our problem is usually highly non-convex. We define tangent vectors as the change in the output with a directional derivative of $f(x;\theta)$ in the direction of $gx_i = \frac{\partial f(x;\theta)}{\partial \theta} \big|_{x=x_i}$: $(D_{gx_i}f)(\theta)=\frac{d}{dt}[f(x;\theta+tgx_i)]|_{t=0}$. We will refer $\nabla: \mathbb{R}^{d_{in}} \rightarrow \mathbb{R}^{N_{\theta}}$ as tangent mapping of input at $\theta$: $\nabla_\theta f(x;\theta) := \frac{\partial{f(x;\theta)}}{\partial {\theta}}$. An intuitive interpretation is that $\nabla$ gives the direction where the parameter vector $\theta$ should be changed to fit best the example $x$. In case of batch learning, at every iteration we estimate the change in $\theta$ with three, for us significant steps: mapping elements of the batch to tangent vectors based on the loss, compute the direction of steepest descent per element and take the mean of the directions to approximate the expected direction, e.g. for first order gradient descent without regularization the update step is at time $t$: $\theta^{t+1} = \theta^{t} + \eta \frac{1}{m} \sum_{i=1}^m \frac{\partial l(f(x;\theta),y)}{\partial \theta} \big|_{\theta=\theta^{t},x=x_i,y=y_i}$ where $\eta \in \mathbb{R}_+$ and $\{(x_1,y_1),..,(x_m,y_m)\}$ is the batch. 

\section{Tangent space sensitivity}
\label{sec:tss}

According to the literature \cite{goodfellow2014explaining,carlini2017adversarial,werpachowski2019detecting}, there are several ways to generate adversarial (or generate smooth augmented) samples with some common assumptions e.g. a generated example should lie in the neighbourhood of a known example or the label of the generated example will be the same as the known example it is close to. The latter presumption will not be in our interest however we will investigate how the tangent map varies if a new example is generated in the vicinity of a known data point. Let $\phi: \mathbb{R}^{d_{in}} \rightarrow \mathbb{R}^{d_{in}}$ be an adversarial generator with a norm e.g. $l2$, $max$ or AEG \cite{werpachowski2019detecting} thus we can assume that $\left \| x - \phi(x) \right \|_{p} \leq \rho$ for  some norm $p$ almost surely. Let us consider the $l2$ norm and an infinitesimal Gaussian perturbation around $x$ with $\delta(x) = \left \| x - \phi(x)\right \|_{2} \sim \mathcal{N}(0,\sigma\mathbf{I})$ as a generator. Thus the expected change in the tangent mapping ($\nabla_\theta f(x;\theta) = \frac{\partial{f(x;\theta)}}{\partial {\theta}}$) will be for some $\sigma<\infty$

\begin{align*}
\nonumber
\mathbf{E}_{\delta(x)} [\left \| \nabla_\theta f(x;\theta)  - \nabla_\theta f(\phi(x);\theta) \right \|_2^2 ] \sim \mathbf{E}_{\delta(x)} \Big[\left \| \frac{\partial{\nabla_\theta f(x;\theta)}}{\partial{x}} \delta(x)\right \|_2^2 \Big ] \leq \sigma \left \| \frac{\partial{\nabla_\theta f(x;\theta)}}{\partial{x}}\right \|_2^2
\end{align*}

where $x \sim D$. The expected change is not directly computable since it varies by input, however we can approximate this connection with an expectation over $D$: $\mathbf{E}_{x \sim D} [\sigma \left \| \frac{\partial{\nabla_\theta f(x;\theta)}}{\partial{x}} \big |_x \right \|_2^2] $. Before we arrive at computationally feasible measures let us define a matrix based on the input variables and a parameter configuration.

\begin{definition}
\textit{Tangent sample sensitivity} of a parametric, smooth feed-forward network $f$ with output in $\mathbb{R}^{d_{out}}$ at input $x \in \mathbb{R}^{d_{in}}$ is a $N_{\theta} \times d_{in}$ dimensional matrix, $Sens_{tan}(x;\theta) := \frac{\nabla_\theta f(x;\theta)|_{\theta}}{\partial x} \big |_{x}= \frac{\partial^2{f(x;\theta)}}{\partial{\theta} \partial{x}} \big |_{\theta,x}$. We define \textit{tangent sensitivity} as the expectation of \textit{tangent sample sensitivity}: $Sens_{tan}(\theta)=\mathbf{E}_{x \sim D}[Sens_{tan}(x;\theta)]$.
\end{definition}

The elements of these matrices represent connections between the input and the network parameters. The entries in the matrix decompose the directed paths along the weights based on the source of the path. A particular element of \textit{tangent sample sensitivity} is a summation over the input-output paths containing the weight parameter with the derivatives of the activation functions according to the position of the weight parameter (for more details, see Appendix~\ref{sec:app_a}): for ReLU networks $Sens_{tan}(x;\theta)_{i,j} = \sum_{path \in P_{i,*,j}^+(x;\theta)} \Pi_{w_l \in path, w_l  \neq w_j} w_l$ where we denote active paths including $w_{j}$ between the $i$-th input node and any output node with $P_{i,*,j}^+(x;\theta) = \cup_{l=\{1,..,d_{out}\}} \{P_{i,l}(x;\theta)| w_j \in P_{i,l}(x;\theta), \forall h_{p_{i,l}}(x;\theta) >0\}$ for an input $x$. Our first bound is independent of input and depends only on the weight parameters.

\begin{theorem}
\label{theo1}
For a biasless feed-forward ReLU network with $k$ layers, $N_{max} = \max_i N_i$, $w_{max}=\max_{w \in \theta} |w|$ and $w_{max_i}=\max_{w \in \theta_i} |w|>0$ for all $i$, the Frobenius norm of \textit{tangent sensitivity} is upper bounded by a $2(k-1)$ degree homogeneous function in $\theta$ as 
\begin{align}
\left \|Sens_{tan}(\theta)\right \|_{F}^2&=\mathbf{E}_{x \sim D}[\left \|Sens_{tan}(x;\theta)\right \|_{F}^2] \nonumber \\
&\leq N_{\theta} d_{in} (N_{max})^{2(k-1)} (\frac{1}{\min_i w_{max_i}}\Pi_{i=1}^k w_{max_i})^2 \nonumber \\
&\leq N_{\theta} d_{in} (N_{max})^{2(k-1)} (w_{max})^{2(k-1)}. \label{bound1}
\end{align}
\end{theorem}

Theorem \ref{theo1} is proved in Appendix~\ref{sec:app_a}. Note, the bound almost never occurs. Both the maximal path count and the uniform maximal weighted paths are very specific cases, when every layer has the same size and the weights are equal. The bound suggest to minimize the $l_{\infty}$ norm over parameters per layer. Our bound coincides with \cite{neyshabur2014search} where the authors suggest layer-wise regularization and consider $l_{\infty}$ for the incoming weights per hidden unit. Max-norm regularization was shown to provide good performance in \cite{srivastava2014dropout}. On the other hand one of the most commonly used regularizer methods is weight decay \cite{krogh1992simple}. It was shown that for ReLU networks per-unit $l_2$ regularization could be very effective because of the positive homogeneity property of ReLU activations. Additionally, the weights can be rescaled in a fashion that all hidden units have similar norm thus the regularizer does not focus on extreme weights. This property suggests that in ReLU networks per-unit $l_2$ regularization may lead comparable results to our norm per layer. In summary, our bound is in accordance with the most common regularization methods over the network parameters. However \textit{tangent sensitivity} may include additional knowledge about the structure and paths inside the network similarly to Path-SGD \cite{NIPS2015_5797}. 

\begin{figure}
\centering
\begin{subfigure}{.4\textwidth}
  \centering
  \includegraphics[width=1\linewidth]{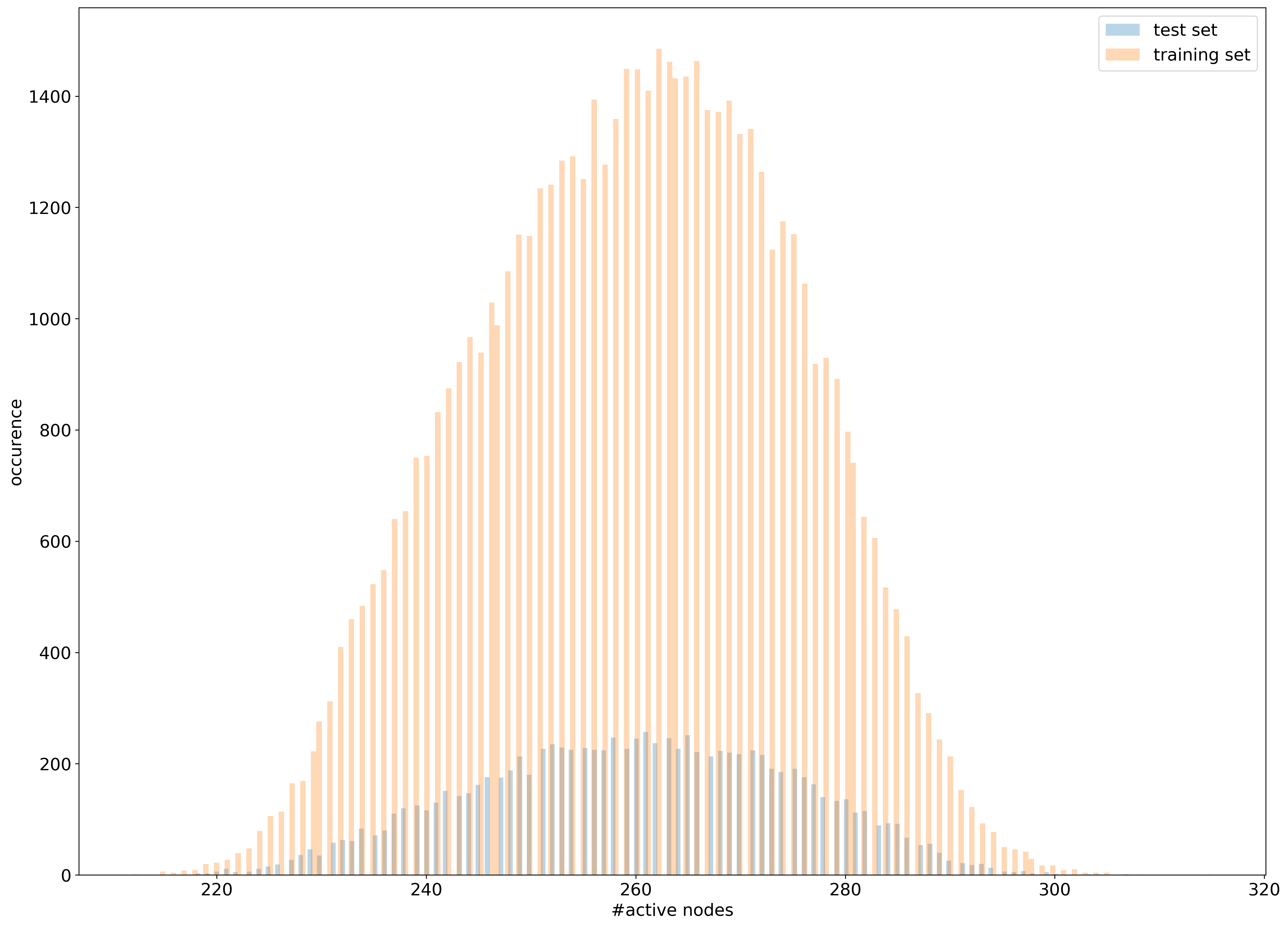}
  \caption{MNIST \cite{lecun1998gradient}}
  \label{fig:sfig11}
\end{subfigure}
\begin{subfigure}{.4\textwidth}
  \centering
  \includegraphics[width=1\linewidth]{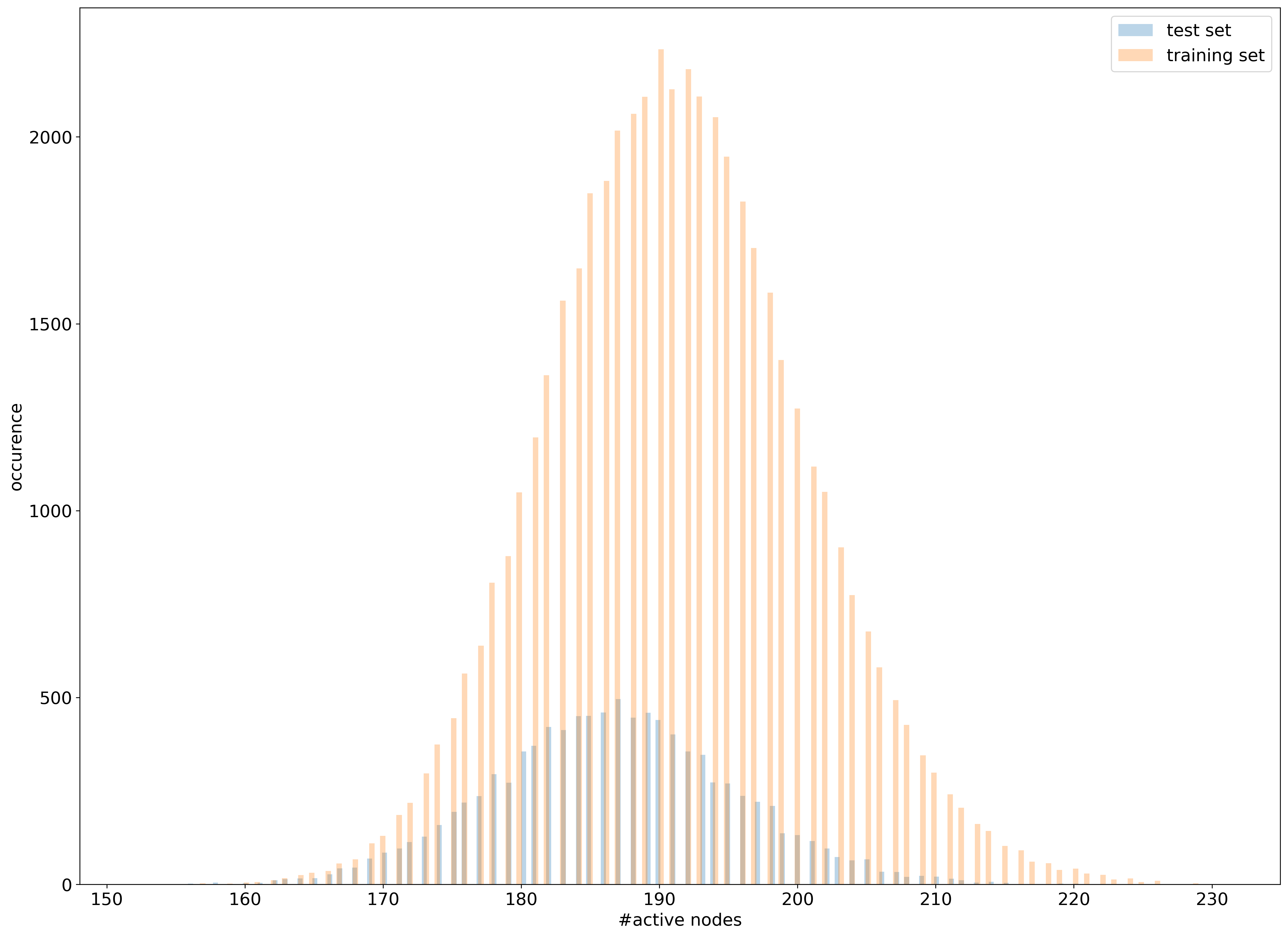}
  \caption{CIFAR-10 \cite{krizhevsky2009learning}}
  \label{fig:sfig12}
\end{subfigure}
\caption{Histogram of active neurons (nodes in the network) after 50 epochs on the training and on the test set in a 4-layer feed-forward ReLU network with $100$ neurons per layer.}
\label{fig:distr_actn}
\end{figure}

We now tighten the bound by discarding the assumption of independence from input and relating it to the structure of paths in the network. The above generic calculation assumes that every path materialized a.k.a. every node has a positive activation along the path. Recent results \cite{hanin2019deep} suggest that we can consider a more realistic case where the number of active nodes is significantly less than the number of nodes in the network, see Fig~\ref{fig:distr_actn}, thus our second bound takes into consideration the distribution of active nodes with the assumption of Gaussian. Let $T(x) = \sum_i n_i(x)$ be the number of active nodes for an input $x$ with $n_i(x)$ as the number of active nodes in the $i$-th layer. 

\begin{theorem}
\label{theo2}
For $x \sim D$ and a biasless feed-forward ReLU network with $w_{max_i}=\max_{w \in \theta_i} |w|$, with the number of active nodes $T(x)$ following a normal distribution $\mathcal{N}(\mu,\sigma)$, the Forbenius norm of \textit{tangent sensitivity} is upper bound by 
\begin{equation}
\label{bound2}
N_{\theta}  d_{in} \sigma^{2(k-1)}\frac{2^{k-1}}{k^{2k}}\frac{(\Gamma(k/2))^2}{\Pi}(\Psi(-(k-1)/2,1/2,-\mu^2/(2\sigma^2))^2 (\Pi_{i=1}^k w_{max_i})^2.
\end{equation}
where $\Psi$ is Krummer's confluent hypergeometric function.  
\end{theorem}

Theorem \ref{theo2} is proved in Appendix~\ref{sec:app_b}. The bound suggests an interesting connection between the depth of the network and the distribution of the number of active nodes. The depth of the network may overcome instability with appropriate weights and mean number of active nodes. In order to probe this connection, we investigate the effect of depth, width, the norm of the weights and the number of active nodes in Section~\ref{sec:results}. Note, we have taken the maximal path count given the number of active units however we do not take into account the distribution of activation patterns with equivalent number of active hidden units. Thereby let us look into the expected sensitivity from a bit different angle to estimate \textit{tangent sensitivity} as we focus on linear regions.

\subsection{Distribution of linear regions}

According to \cite{montufar2014number} the possible number of linear regions with the same activation pattern is $\Big(\prod\limits_{i=1}^{k-1}\floor*{\frac{n_i}{d_{in}}}^{d_{in}}\Big)$ if we consider networks with a single output. In practice the occurrence of every region is extremely rare. The following lemma suggest us to investigate how the regions cover the input space a.k.a. how the volume of convex activation regions induced by the network. 
\begin{lemma}
\label{lemma1}
For each element in an activation region $R(A;\theta)$ \textit{tangent sample sensitivity} is identical. 
\end{lemma}
Lemma \ref{lemma1} is proved in Appendix~\ref{sec:app_c}. Worth to mention, the lemma does not hold for linear regions. To see why, let us consider two neighbouring activation regions which differ only in one active neuron. The number of positive paths for elements in the neighbouring region will be higher if the neuron was not active and lower if the neuron was active in the first region. We refer to \textit{tangent sensitivity} for elements in an activation region with $Sens_{tan}(A;\theta)$. Without loss of generality we may assume that the input space is compact thus every activation region has finite volume as $\mathit{vol}(R(A;\theta))<\infty$ therefore the mean sensitivity for the compact input space is $Sens_{tan}(\theta) = \sum_{A \in \mathit{A}^+} \mathit{vol}(R(A;\theta)) Sens_{tan}(A;\theta)$ where $\mathit{A}^+$ is the finite set of activation patterns of non-empty activation regions. As linear regions are represented by a finite set of linear inequalities their volume is exactly the volume of the bounding polytope. In case of activation regions these polytopes are convex. Unfortunately, computing the volume of an explicit polytope is $\#P$-hard thus infeasible. In \cite{lovasz2006simulated} the authors suggested an $O^*(d_{in}^4)$ (without additional terms, $O^*(d_{in}^3)$ in a special case \cite{cousins2016practical}) algorithm for estimating the volume of a single convex body by simulated annealing. In a recent result \cite{chakrabarti2019quantum}, complexity was further improved with quantum oracles to $O^*(d_{in}^{3})$, but the computations remain too expensive for tasks where neural networks have an advantage (e.g. high dimensional input dimension). It is worth noting that the convex body structure of a ReLU network is a well-defined subset, a hyperplane arrangement therefore we are interested in the volume of many convex polytopes. 

There are several ways to relax volume computation at the cost of accuracy. For example, authors in \cite{zhang2020empirical} estimated the radius of inspheres of linear regions by finding a point inside the polytope with the largest distance from the closest facet with solving a convex optimization problem. This estimation measures the narrowness of a region and only valid for activation regions. However none of the previously mentioned algorithms take into account the data distribution even if there are large regions without any support. Straightaway empirical estimation of the expected sensitivity may be difficult as by Hoeffding's inequality \cite{hoeffding1994probability} $Pr_{x \sim D}\{|\frac{1}{T}\sum_i^T\left \|Sens_{tan}(x_i;\theta)\right \|_{F}^2-\left \|Sens_{tan}(\theta)\right \|_{F}^2|>\epsilon\} \leq \exp(-\frac{1}{2}\frac{\epsilon^2 T}{(Sens_{max,F})^2})$ can be meaningless if the maximal Frobenius norm of \textit{tangent sample sensitivity}, $Sens_{max,F}=\max_x \left \| Sens_{tan}(x;\theta) \right \|_F^2$ is high. Note, the maximal sensitivity for a particular network is finite. Based on Lemma~\ref{lemma1} \textit{tangent sensitivity} depends on how the activation patterns distribute over the compact input space as 
\begin{equation}
\label{b3}
\mathbf{E}_{x \sim D}[\left \|Sens_{tan}(x;\theta)\right \|_{F}^2] =\mathbf{E}_{A \sim p(A;x,\theta)}[\left \|Sens_{tan}(A)\right \|_{F}^2]
\end{equation} 
where we denote the probability of an input is in an activation region with $p(A;x,\theta)$. The question remains: how can we determine the probability of a region without exactly computing the volume? As the number of activation regions may be larger than the size of an available dataset, practical calculation of relative frequency could be misleading. To overcome this we suggest to relaxation. By shallow networks independence of hidden units may seem an acceptable strong assumption, but in deep networks this is no longer the case thus by assuming Markov property inside the network $p(A_{i,j};x,\theta) = p(A_{i,j}| \{h_{i-1,1}(x;\theta),...,h_{i-1,N_{i-1}}(x;\theta)\},\theta) \approx p(A_{i,j}| h_{i,j}(x;\theta))$. As expected the activation of a neuron depends on the preactivation of the neuron. Now, let us assume that for the $l$-th hidden unit (the $j$-th unit in the $i$-th layer) $\log\frac{p(A_{l}=1|x;\theta)}{1-p(A_{l}=1|x;\theta)} \approx h_{l}(x;\theta)$ than for an input $x$ and an activation pattern $A$ the approximated probability is   
\begin{equation}
\label{est_p}
p(A=\{a_{l};l\in \{1,2,..,N\}\}|x;\theta) \approx \Pi_{l|a_{l}=1} \sigma(h_{l}(x;\theta)) \Pi_{l | a_{l}=-1} (1-\sigma(h_{l}(x;\theta))
\end{equation}
where $\sigma(z) = 1/(1+exp(-z))$ denotes the sigmoid function. Note that this approximation is closely related to the margin distribution \cite{jiang2018predicting} as the margins increase with the above probability inside an activation region for the active nodes and decreases outside (for more detail see Appendix~\ref{sec:app_c}). 


\section{Tangent space sensitivity and generalization}
\label{sec:tss_gen}

The authors in \cite{novak2018sensitivity} established that fully trained (trained till zero error on the training set) neural networks show significantly more robust behaviour in the vicinity of the training data manifold, especially with random labels, in comparison to other subsets of the input space. They measure robustness on the training set by sampling around the training points and compute the Jacobian of the function realized by the network w.r.t the input. In comparison we would like to use our previously derived measures without exactly calculating the derivatives and estimate the empirical sensitivity on the training and on the test set. According to Lemma~\ref{lemma1} inside an activation region \textit{tangent sample sensitivity} is constant thus the volume of regions determine global stability. We may introduce some practical estimations of loss on the test set ($X_{te}$) based on various sensitivity measures and the loss on the training set ($X_{tr}$ with the corresponding target $Y_{tr}$):

\begin{itemize}
\item Layer-wise norm sensitivity, $Sens_{tan}^1$ (eq.~\ref{bound1}): the bound does not depend on input however the layer-wise $l_{\infty}$ norm of the parameters change throughout learning therefore we may estimate the loss at time t (learning step t) based on the inverse change in maximal sensitivity (eq.~\ref{bound1}) and loss measured on the training set at time t:
\begin{equation}
\label{diff:d_b1}
\hat{l}^1(X_{te};\theta^{(t)}) := \frac{Sens_{tan}^1(\theta^{(t-1)}))}{Sens_{tan}^1(\theta^{(t)})} l({X_{tr},Y_{tr};\theta^{(t)}}) = \frac{(\Pi_{i=1}^k w_{max_i}^{(t-1)})^2}{(\Pi_{i=1}^k w_{max_i}^{(t)})^2} l({X_{tr},Y_{tr};\theta^{(t)}})
\end{equation}
where $w_{max_i}^{(t)}$ is $l_{\infty}$ norm in the $i$-th layer at time t. The missing parts of (eq.~\ref{bound1}) are invariable throughout learning.

\item Maximal sensitivity, $Sens_{tan}^2$ (eq.~\ref{bound2}): similarly, we may estimate test loss based on the distribution of the number of active nodes: 
\begin{equation}
\label{diff:d_b2}
\hat{l}^2(X_{te};\theta) := \frac{\psi^*(k,\mu_{te},\sigma_{te})}{\psi^*(k,\mu_{tr},\sigma_{tr})}  l(X_{tr},Y_{tr};\theta)
\end{equation}
where $\psi^*(k,\mu,\sigma) = \sigma^{2(k-1)} \Psi(-(k-1)/2,1/2,-\mu^2/(2\sigma^2))^2$ and the corresponding normal distributions are $\mathcal{N}(\mu_{tr},\sigma_{tr})$ and $\mathcal{N}(\mu_{te},\sigma_{te})$ for the training and the test set respectively. Note, the missing parts of (eq.~\ref{bound2}) are invariable if the graph of the network is fixed. At any state of the network the difference between the estimated sensitivity is based only on how the distribution of the number of active neurons differ in the two sets and the depth of the network while concealing the difference in activation patterns given the sets. 

\item Empirical sensitivity, $Sens_{tan}^3$ (eq.~\ref{b3}): assuming the empirical estimation of $p(A;\theta)$ in (eq.~\ref{est_p}) and (eq.~\ref{bound2}) we define empirical \textit{tangent sensitivity} as $\hat{S}(X;\theta) = \sum_{A_i \in \mathit{A}^+} p_{x \in X}(A_i;\theta) Sens_{tan}(A_i;\theta)$. The corresponding estimation of test loss:
\begin{equation}
\label{diff:d_pr}
\hat{l}^3(X_{te};\theta) :=  \frac{\hat{S}(X_{te};\theta)}{\hat{S}(X_{tr};\theta)}  l(X_{tr},Y_{tr};\theta). 
\end{equation}
\end{itemize}

In addition to distribution of activation regions the above estimations coincide with previous results that connect generalization to the norm of network parameters \cite{bartlett2017spectrally}, to maximal capacity \cite{lin2016does,neyshabur2017exploring,cao2019generalization} and to margin distribution \cite{jiang2018predicting}. 
  
\begin{figure}
\centering
\begin{subfigure}{.45\textwidth}
  \centering
 \includegraphics[width=1\linewidth]{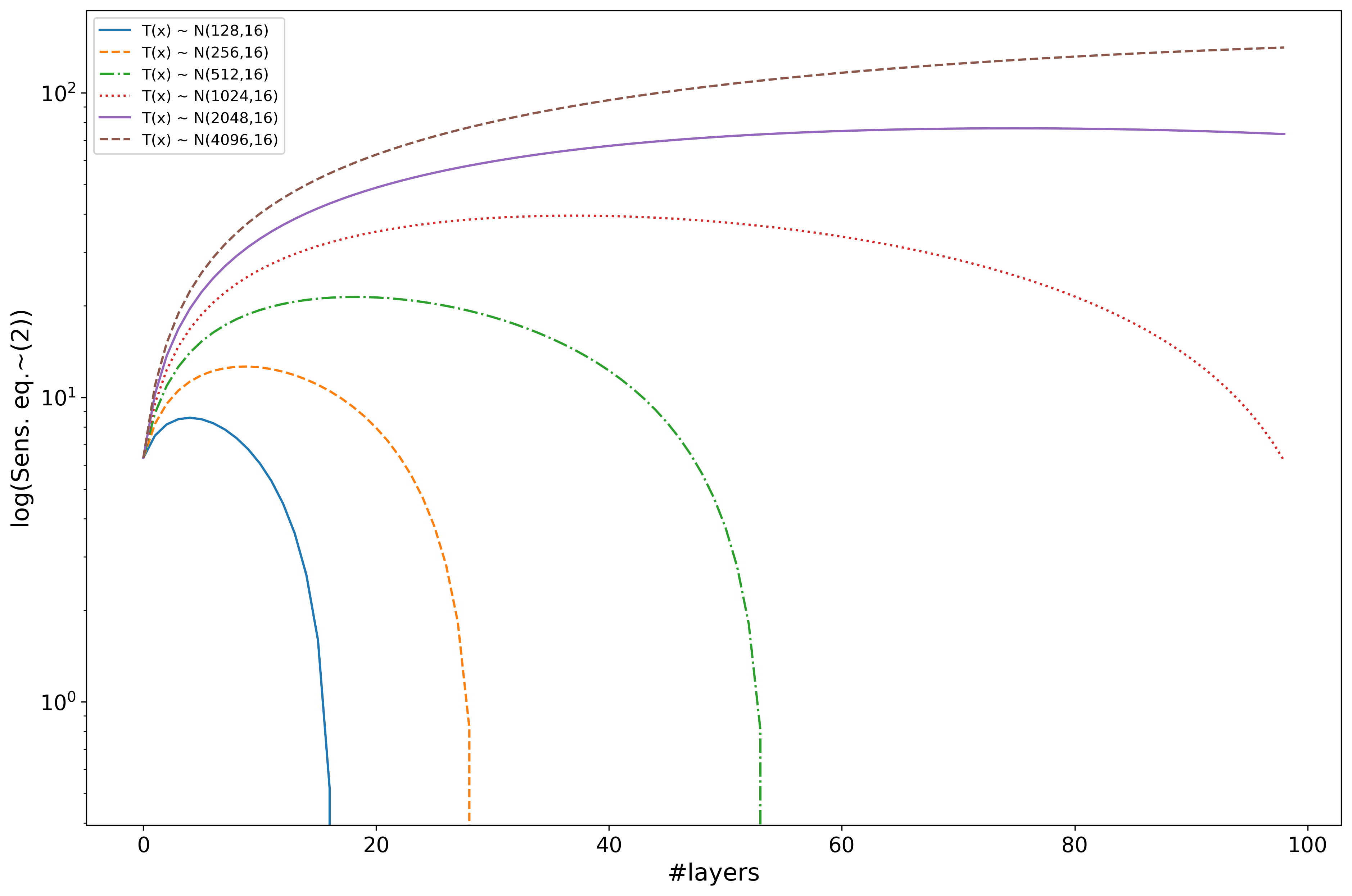}
  \caption{$N_{layer}=1k$,$w_{max}=0.1$, log-scale}
  \label{fig:sfig7}
\end{subfigure}%
\begin{subfigure}{.45\textwidth}
  \centering
  \includegraphics[width=1\linewidth]{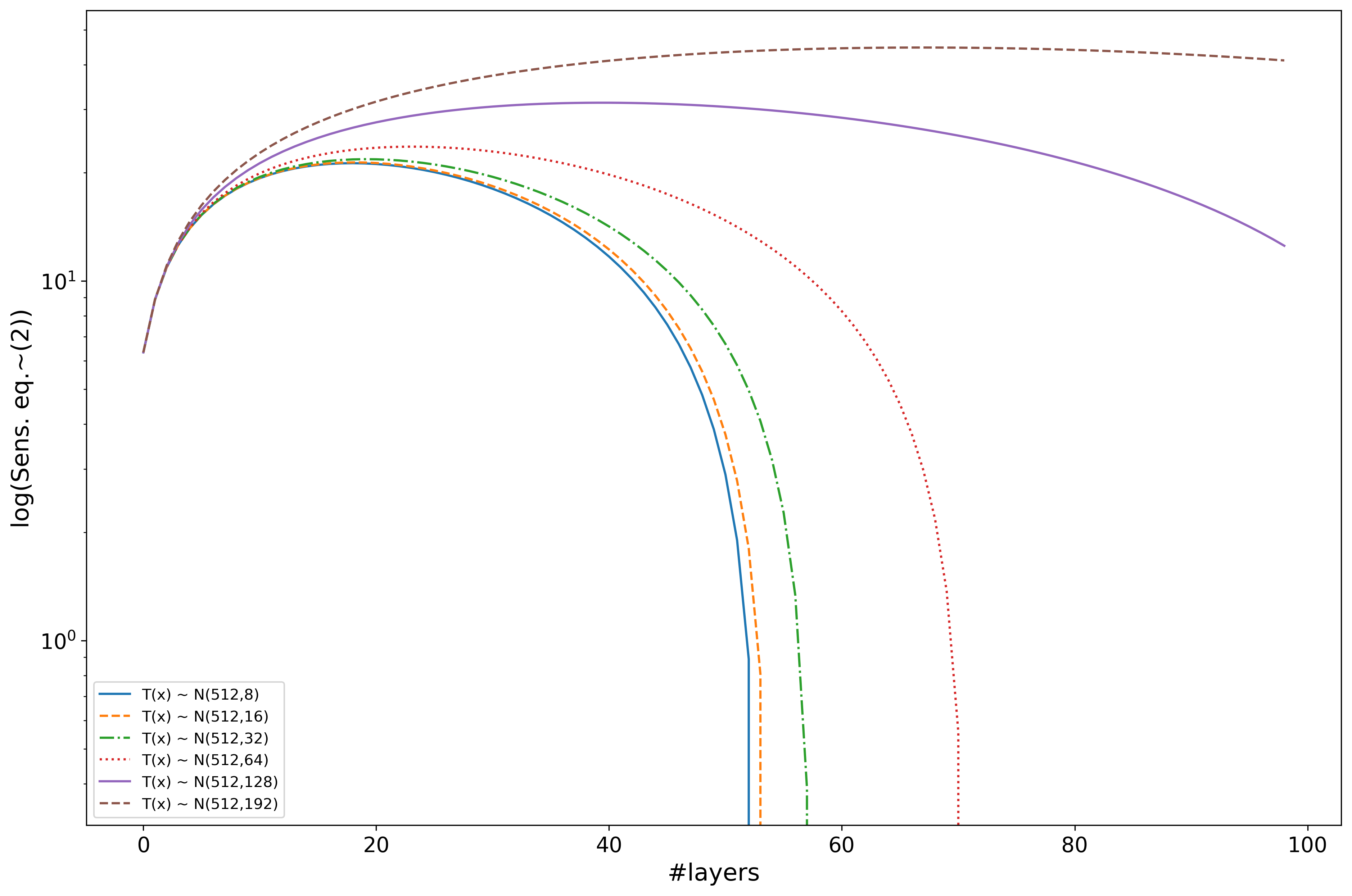}
  \caption{$N_{layer}=1k$,$w_{max}=0.1$, log-scale}
  \label{fig:sfig8}
\end{subfigure}

\begin{subfigure}{.45\textwidth}
  \centering
 \includegraphics[width=1\linewidth]{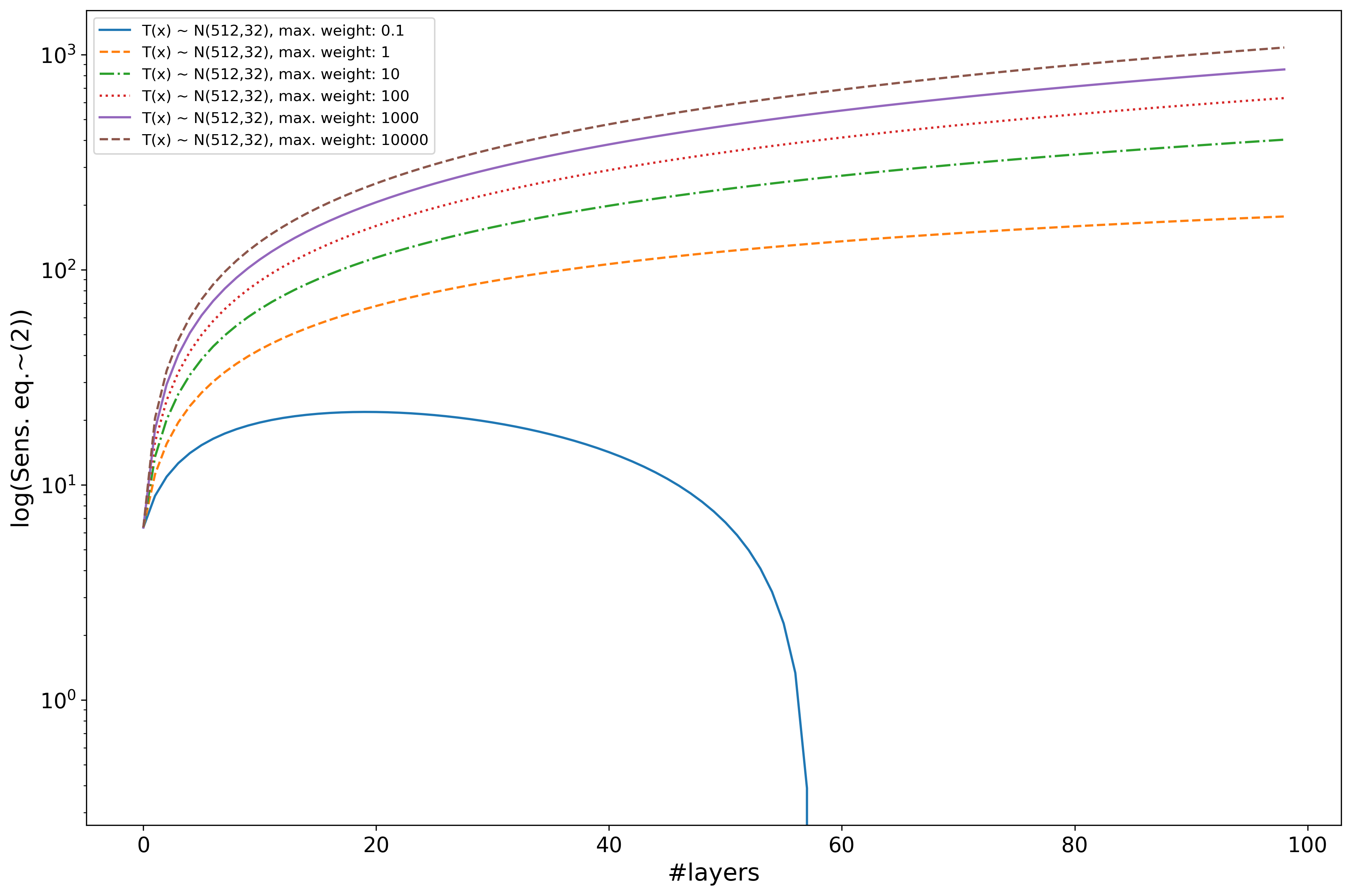}
  \caption{$N_{layer}=1k$, log-scale}
  \label{fig:sfig14}
\end{subfigure}%
\begin{subfigure}{.45\textwidth}
  \centering
  \includegraphics[width=1\linewidth]{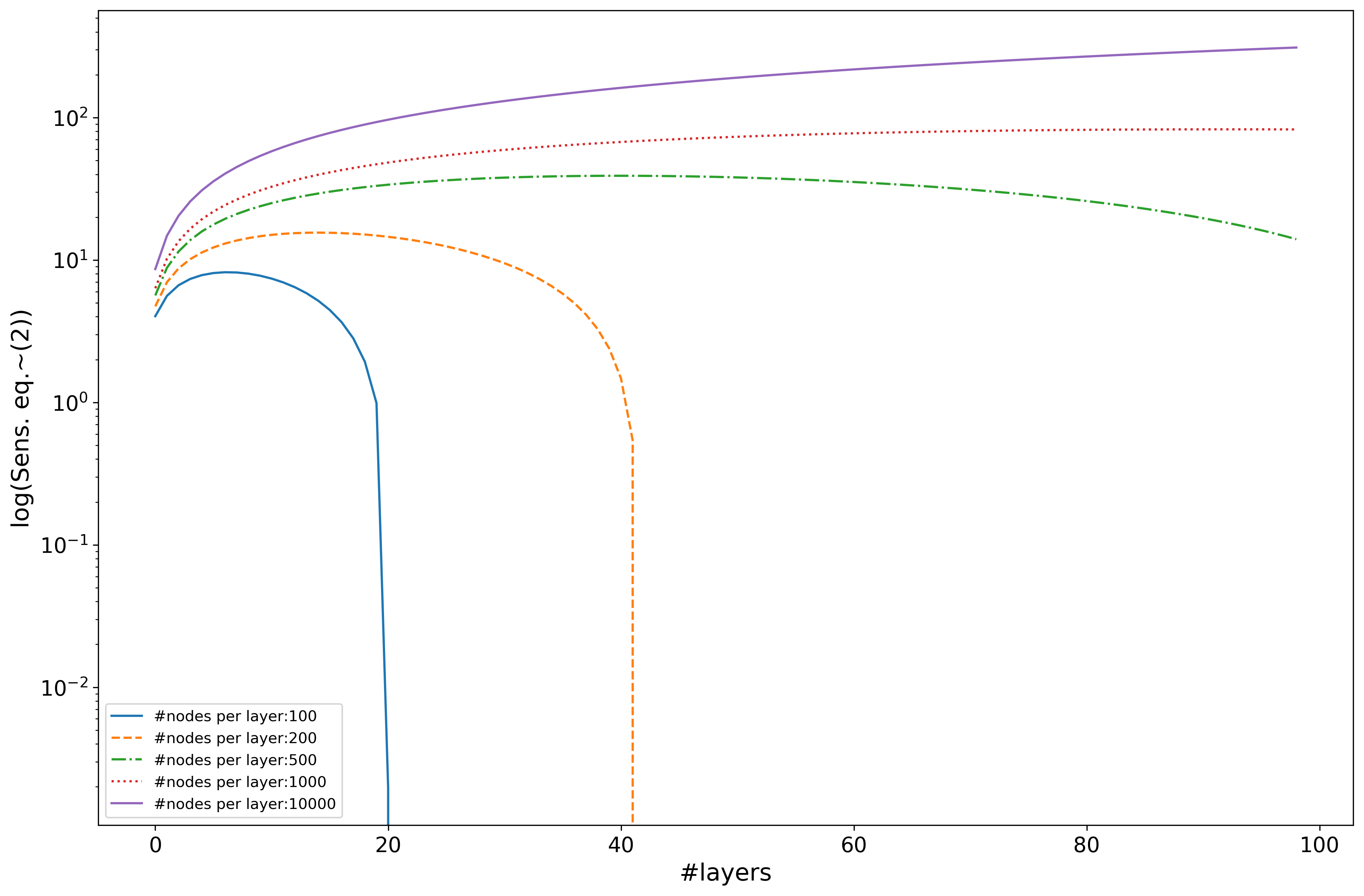}
  \caption{$\mu=0.48N_{\theta}$, $\sigma=0.04N_{\theta}$ (\ref{fig:sfig12}), log-scale}
  \label{fig:sfig15}
\end{subfigure}

\caption{Change in upper bound (eq.~\ref{bound2}) induced by mean (\ref{fig:sfig7}), variance (\ref{fig:sfig8}), maximal norm (\ref{fig:sfig14}) and width of layers (\ref{fig:sfig8}).}
\label{fig:synth}
\end{figure}

\section{Experiments}
\label{sec:results}

Building upon the discussion in Sections~\ref{sec:tss} and ~\ref{sec:tss_gen}, we experiment with feed-forward fully-connected ReLU networks. We implemented the measures in Section~\ref{sec:tss_gen}. Based on the analysis of (eq.~\ref{bound1}) and (eq.~\ref{bound2}) we observed that the most important factor in sensitivity is the depth of the network followed by the layer-wise norm of the parameters and the empirical variance of number of active neurons in the network, for details see Fig.~\ref{fig:synth}. Additionally, we found that the upper bound of \textit{tangent sensitivity} with proper regularization (e.g. low norm) after reaching a certain depth starts to decrease, supporting one of the fundamental phenomenon of deep learning, deeper networks may generalize better. 

\begin{figure}
\centering

\begin{subfigure}{.45\textwidth}
  \centering
  \includegraphics[width=1\linewidth]{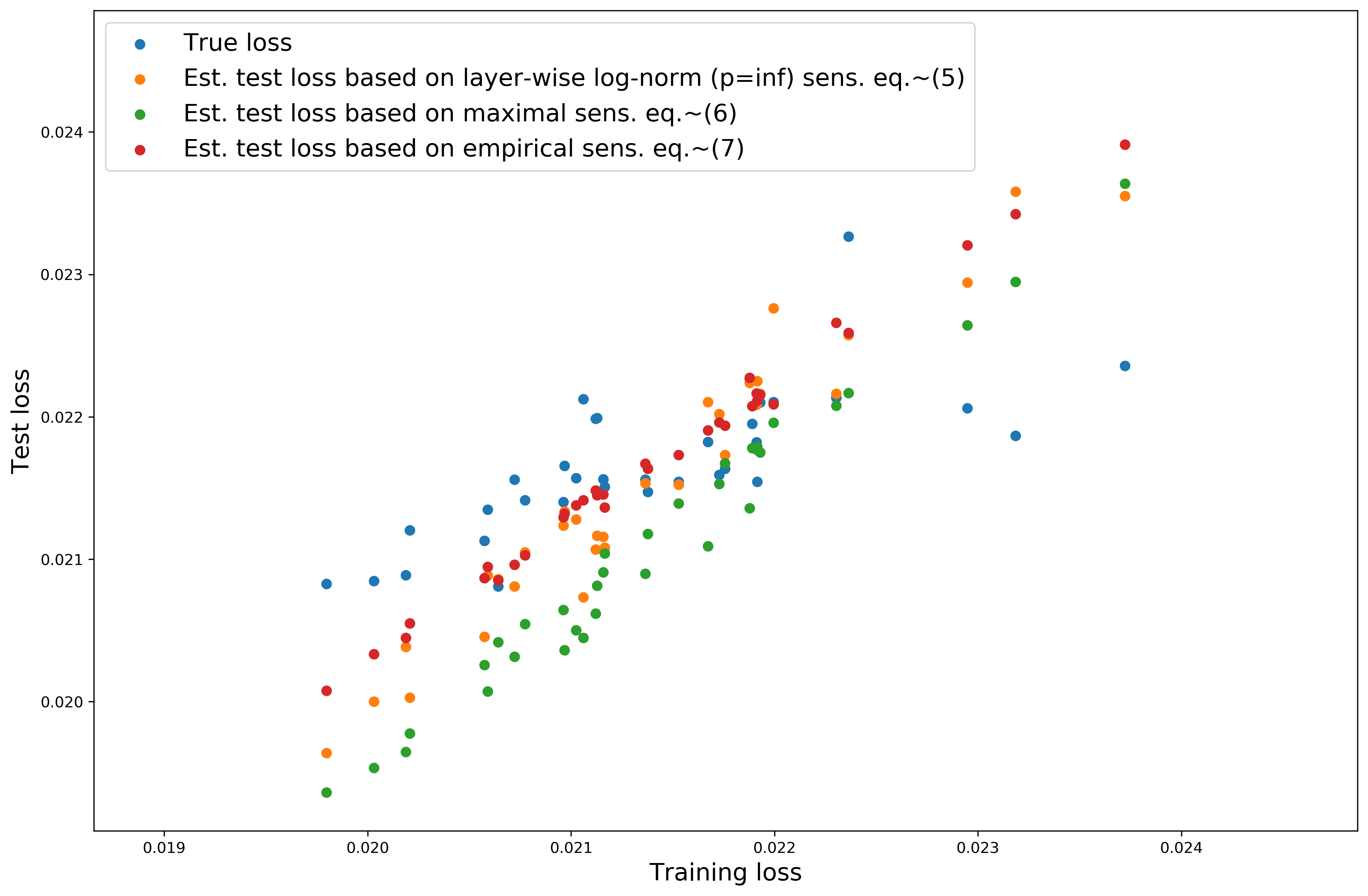}
  \caption{Estimated test cross-entropy loss.}
  \label{fig:sfig2}
\end{subfigure}
\begin{subfigure}{.45\textwidth}
  \centering
 \includegraphics[width=1\linewidth]{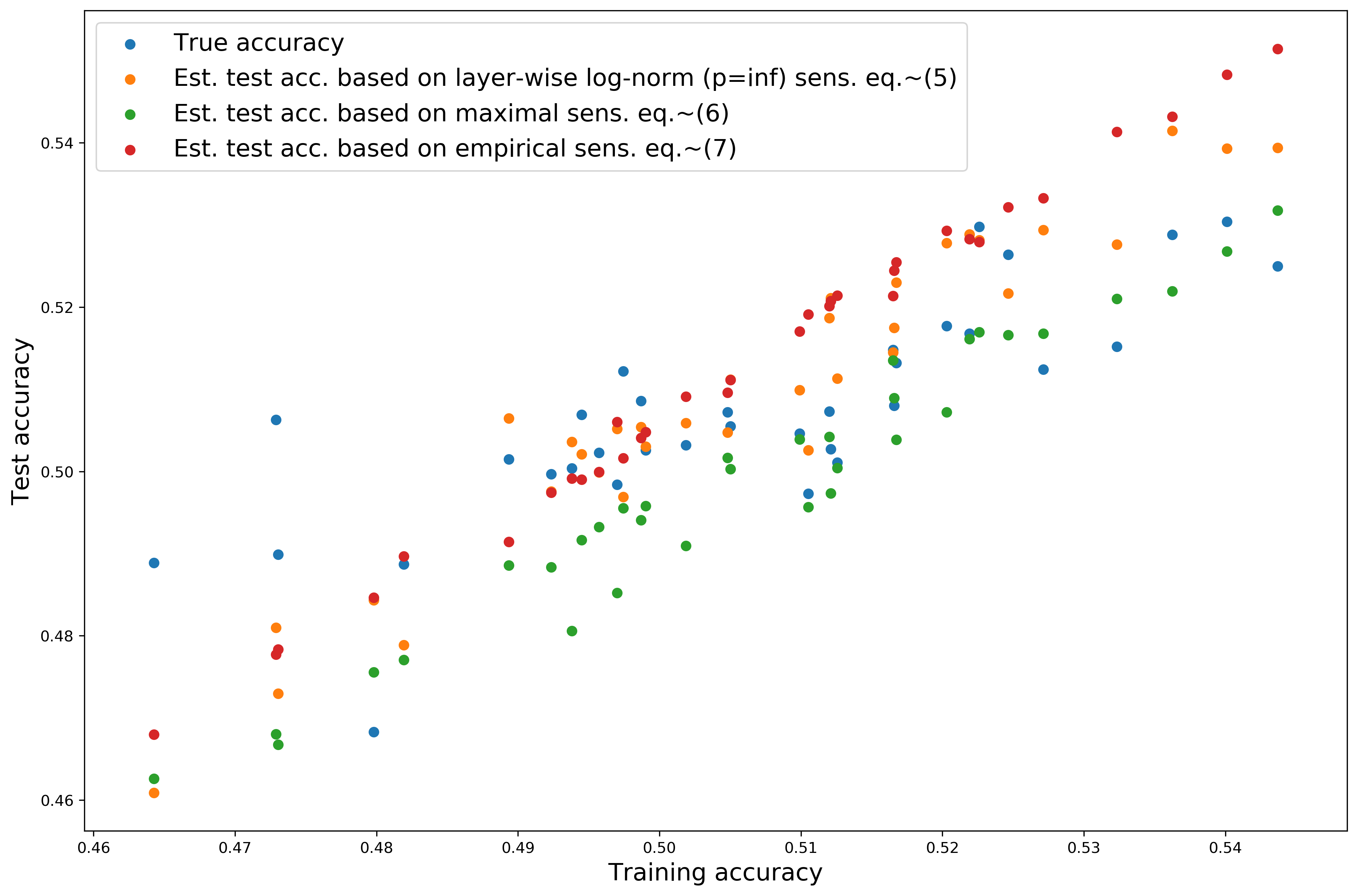}
  \caption{Estimated test accuracy.}
  \label{fig:sfig4}
\end{subfigure}

\caption{Estimated test cross-entropy loss (\ref{fig:sfig2}) and test accuracy (\ref{fig:sfig4}) on CIFAR-10 dataset with ReLU network on different states (trained with stochastic gradient descent with batch size of $64$, learning rate with $\alpha=0.05$ and weight decay with $\beta=0.0005$) based on layer-wise log-norm ($l_{\infty}) $ (eq.~\ref{diff:d_b1}), maximal (eq.~\ref{diff:d_b2}) and empirical \textit{tangent sensitivity} (eq.~\ref{diff:d_pr}).}
\label{fig:corr}
\end{figure}

Furthermore, we investigated how empirical accuracy and cross-entropy loss related to \textit{tangent sensitivity} in case of feed-forward fully-connected ReLU networks with four hidden layers on the CIFAR-10 \cite{krizhevsky2009learning} dataset. During the experiments we used the five training batches of CIFAR-10 as training set and the test batch as test set. The network parameters were optimized with stochastic gradient descent with weight decay. Results in Fig.~\ref{fig:corr} show that our previously introduced measures may estimate change in empirical generalization gap at some extent. We found that the upper bound of \textit{tangent sensitivity} may indicate exponentially large change in loss because of the layer-wise $l_{\infty}$ norm of the parameters thus we modified our estimation by taking the logarithm of sensitivity instead of simply taking (eq.~\ref{bound1}) in (eq.~\ref{diff:d_b1}). All estimations performed very similarly. The lowest Mean Absolute Error (MAE) for cross-entropy loss and accuracy were achieved by empirical sensitivity (eq.\ref{diff:d_pr}) and layer-wise log-norm sensitivity (eq.~\ref{diff:d_b1}) respectively. Additional experiments with detailed description about the experiments and the implementation can be found in Appendix~\ref{sec:app_d}.




\section{Conclusions}

In this paper we proposed measures of sensitivity to adversarial perturbations in feed-forward neural networks without considering any label. To calculate the sensitivity of the network, we estimate the change to small perturbations in the tangent vectors with taking the derivative of the tangent vectors w.r.t the input. We found that \textit{tangent sensitivity} in ReLU networks is related to the number of active paths between input-output pairs and the norm of the weight parameters. We also found that \textit{tangent sensitivity} is constant inside activation regions and the expected sensitivity is related to the distribution of the activation regions. We plan to examine this connection in future work e.g. convergence of distribution of activation regions during learning and generalize the results to activation functions with bounded first derivatives. In addition, our initial assumptions merit further investigation of residual, convolutional and recurrent network structures together with autoencoders. Furthermore, our work was limited to smooth transformations in input omitting important non smooth augmentation methods e.g. image mirroring. A natural next step would be to connect \textit{tangent sensitivity} with information geometry as feed-forward neural networks usually has a Riemannian metric structure \cite{amari1996neural,liang2017fisher} and examine how generalization induced by the differential structure while constructing regularization methods to minimize sensitivity, suggesting non-trivial network structures and exploiting invariance properties of Fisher information \cite{ay2017information} among others.




\newpage

\section*{Broader Impact}

Our work focus on the stability of neural networks from a theoretical perspective, concentrating on one of the most widely used network structure, the feed-forward fully connected ReLU networks. As such, our findings may help to develop more robust network structures and optimization methods as a possible outcome. In case if our results help to build newly developed models with increased generalization we cannot ignore the possibility that someone may build a model not in the best interest of everyone.



\small
\bibliographystyle{plain}
\bibliography{tangent}

\section{Appendix A}
\label{sec:app_a}

In this section we prove Theorem 3.1. Elements of the \textit{tangent sample sensitivity} matrix represent connections between the input variables and the network parameters. The entries in the \textit{tangent sample sensitivity} matrix decompose the directed paths along the weights based on the source of the path. To show the connection, let us first calculate, by symmetry of second derivatives, the derivative w.r.t the $l$-th input variable $x_l$ for biasless fully connected networks with linear activations. For example, for a network with two input nodes, one output node and two hidden nodes in a single hidden layer the derivative will be a simple summation: $\frac{\partial f(x;\theta)}{\partial x_1} = w_{2,1} w_{1,1} + w_{2,2} w_{1,3}$ as the original function is $f(x;\theta) = w_{2,1}(w_{1,1} x_1 + w_{1,2} x_2) + w_{2,2}(w_{1,3} x_1 + w_{1,4} x_2)$. If we increase the number of hidden nodes the summation will have an additional element corresponding to the new node. In comparison, if we increase the number of hidden layers the number of elements in the summation multiply with the width of the new hidden layer, e.g. for an additional hidden layer with two hidden units the corresponding derivative will be  $\frac{\partial f(x;\theta)}{\partial x_1} = w_{3,1} w_{2,1} w_{1,1} + w_{3,1} w_{2,2} w_{1,3}+w_{3,2} w_{2,3} w_{1,1} + w_{3,3} w_{2,2}w_{1,3}$. Observe that each element in the summation corresponds to an existing directed path in the network graph. In addition, the partial derivative w.r.t a network parameter is a summation over the elements including the corresponding weight e.g. in our example$\frac{\partial^2 f(x;\theta)}{\partial x_1 \partial w_{3,1}} = w_{2,1} w_{1,1} + w_{2,2} w_{1,3}$ since out of the four directed paths between the input node and the output node only two contain $w_{3,1}$. If we replace the activations with ReLU activations the elements in the summation including hidden nodes with negative preactivations will be zero. Bias variables may change preactivations but neither increase or decrease the number of paths. Now, we denote \textit{active paths} including $w_{j}$ between the $i$-th input node and any output node with $P_{i,*,j}^+(x;\theta) = \cup_{l=\{1,..,d_{out}\}} \{P_{i,l}(x;\theta)| w_j \in P_{i,l}(x;\theta), \forall h_{p_{i,l}}(x;\theta) >0\}$ for an input $x$ thus we can derive an element of the \textit{tangent sample sensitivity} matrix with a summation over the active paths $Sens_{tan}(x;\theta)_{i,j} = \sum_{P_{i,j}(x)^+} \Pi_{w_l \in P_{i,j}(x)^+, w_l  \neq w_j} w_l$. In our first bound we consider biasless ReLU networks and maximal path counts. 

\begin{customthm}{3.1}\label{theo1_app}
For a biasless feed-forward ReLU network with $k$ layers, input dimension $d_{in}$, $N_{\theta}$ trainable parameters, $N_{max} = \max_i N_i$, $w_{max}=\max_{w \in \theta} |w|$ and $w_{max_i}=\max_{w \in \theta_i} |w|>0$ for all $i$, the Frobenius norm of \textit{tangent sensitivity} is upper bounded by a $2(k-1)$ degree homogeneous function in $\theta$ as 
\begin{align*}
\nonumber
\left \|Sens_{tan}(\theta)\right \|_{F}^2&=\mathbf{E}_{x \sim D}[\left \|Sens_{tan}(x;\theta)\right \|_{F}^2] \nonumber \\
&\leq N_{\theta} d_{in} (N_{max})^{2(k-1)} (\frac{1}{\min_i w_{max_i}}\Pi_{i=1}^k w_{max_i})^2 \nonumber \\
&\leq N_{\theta} d_{in} (N_{max})^{2(k-1)} (w_{max})^{2(k-1)}. 
\end{align*}
\end{customthm}
\begin{proof} 
In a fully connected feedforward network the set of paths between an input and an output node through a specific edge is either empty (the edge is in the first layer but not connected to the input node), $\Pi_{i=2}^k N_i$ (the edge is in the first layer and connected to the input node), $\Pi_{i=1}^{k-1} N_i$ (the edge is in the last layer) or for an intermediate edge between the $j$-th and next layer $\Pi_{i \neq j,i\neq j+1} N_i$ thus the maximal number of paths between any input-output pair will be less than $(N_{max})^{k-1}$ with $N_{max} =\max_i N_i$. Similarly, along a path the maximal factor in a layer is the highest absolute valued weight $w_{max_i}=\max_{w \in \theta_i} |w|$ and the product will be less or equal than the product of maximal absolute weights for any path $\Pi_{w_l \in P_{i,j}(x)^+, w_l  \neq w_j} w_l \leq \frac{1}{\min_i w_{max_i}} \Pi_{i=1}^k w_{max_i}$ as $\forall i$ $w_{max_i}>0$ thus for any input $x$ every element in $Sens_{tan}(x;\theta)$ will be less than $(N_{max})^{k-1} (w_{max})^{k-1}$. As the matrix has $d_{in} \times N_{\theta}$ elements and the Frobenius norm is $\sum_{i,j}^{N_{\theta},d_{in}} Sens_{tan}(x;\theta)_{i,j}^2$ we get the bound. 
\end{proof}

\section{Appendix B}
\label{sec:app_b}

\begin{figure}
\centering
\begin{subfigure}{.4\textwidth}
  \centering
  \includegraphics[width=1\linewidth]{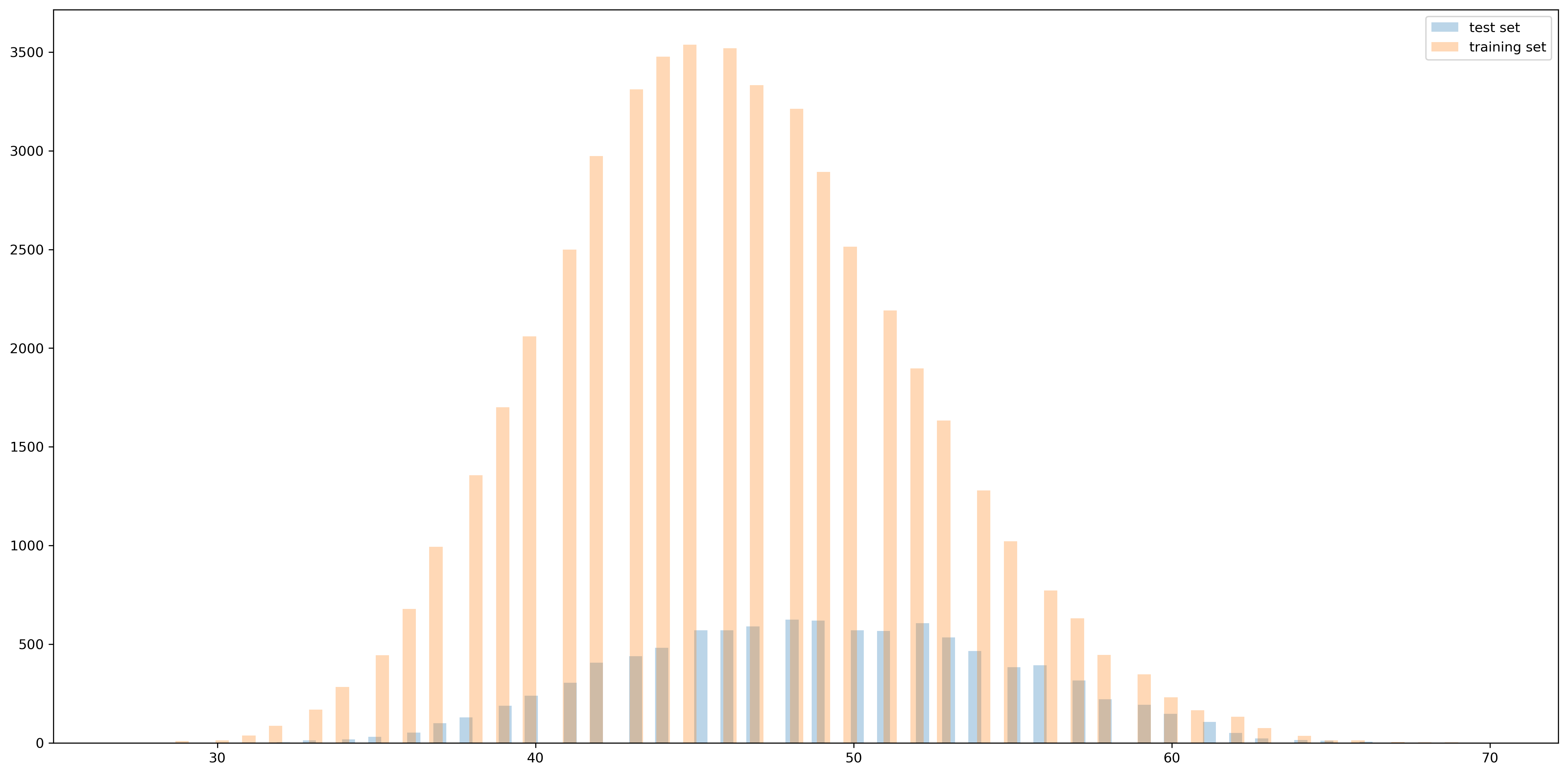}
  \caption{Layer-1}
  \label{fig:sfig51}
\end{subfigure}
\begin{subfigure}{.4\textwidth}
  \centering
  \includegraphics[width=1\linewidth]{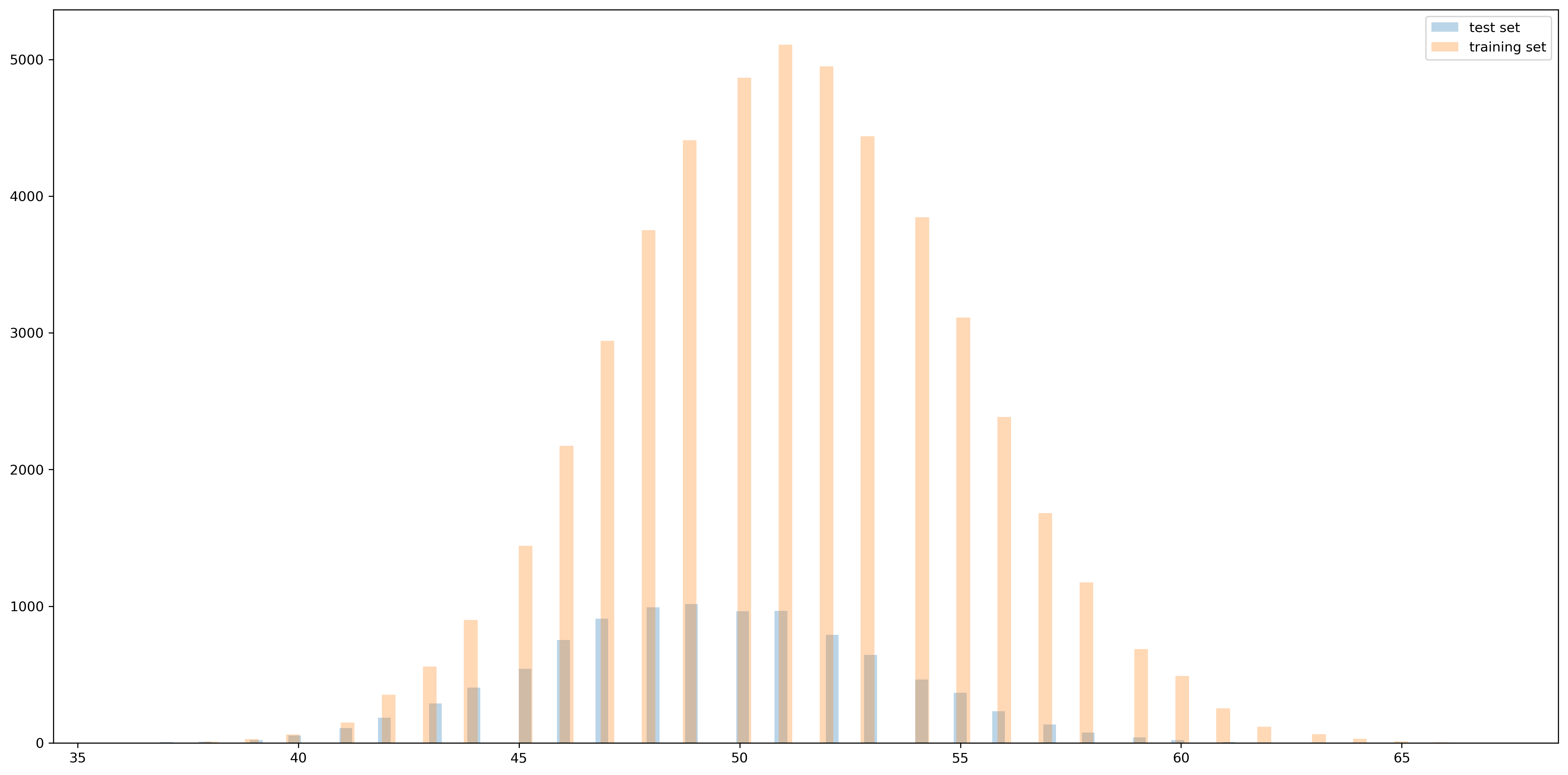}
  \caption{Layer-2}
  \label{fig:sfig52}
\end{subfigure}

\centering
\begin{subfigure}{.4\textwidth}
  \centering
  \includegraphics[width=1\linewidth]{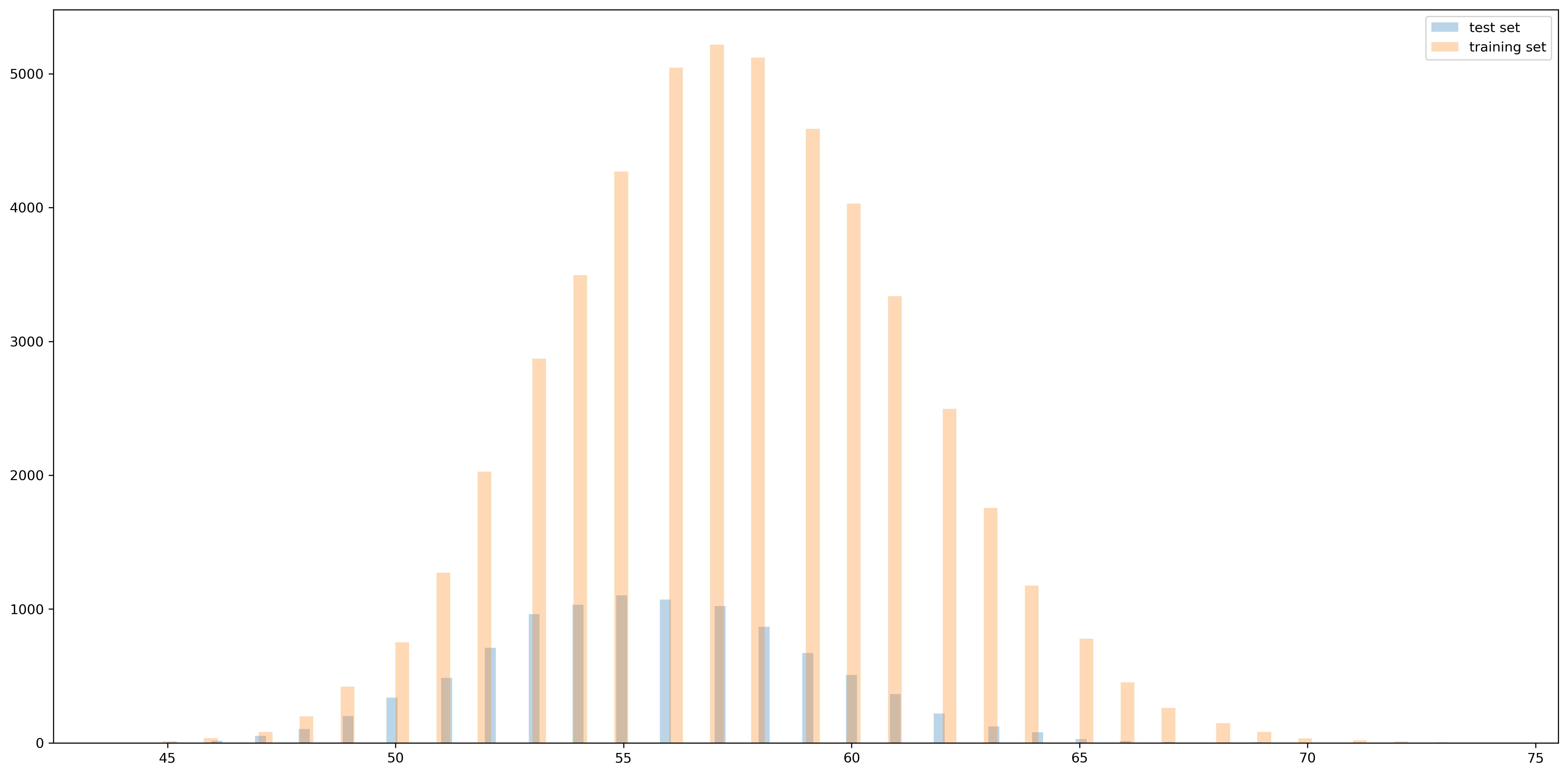}
  \caption{Layer-3}
  \label{fig:sfig53}
\end{subfigure}
\begin{subfigure}{.4\textwidth}
  \centering
  \includegraphics[width=1\linewidth]{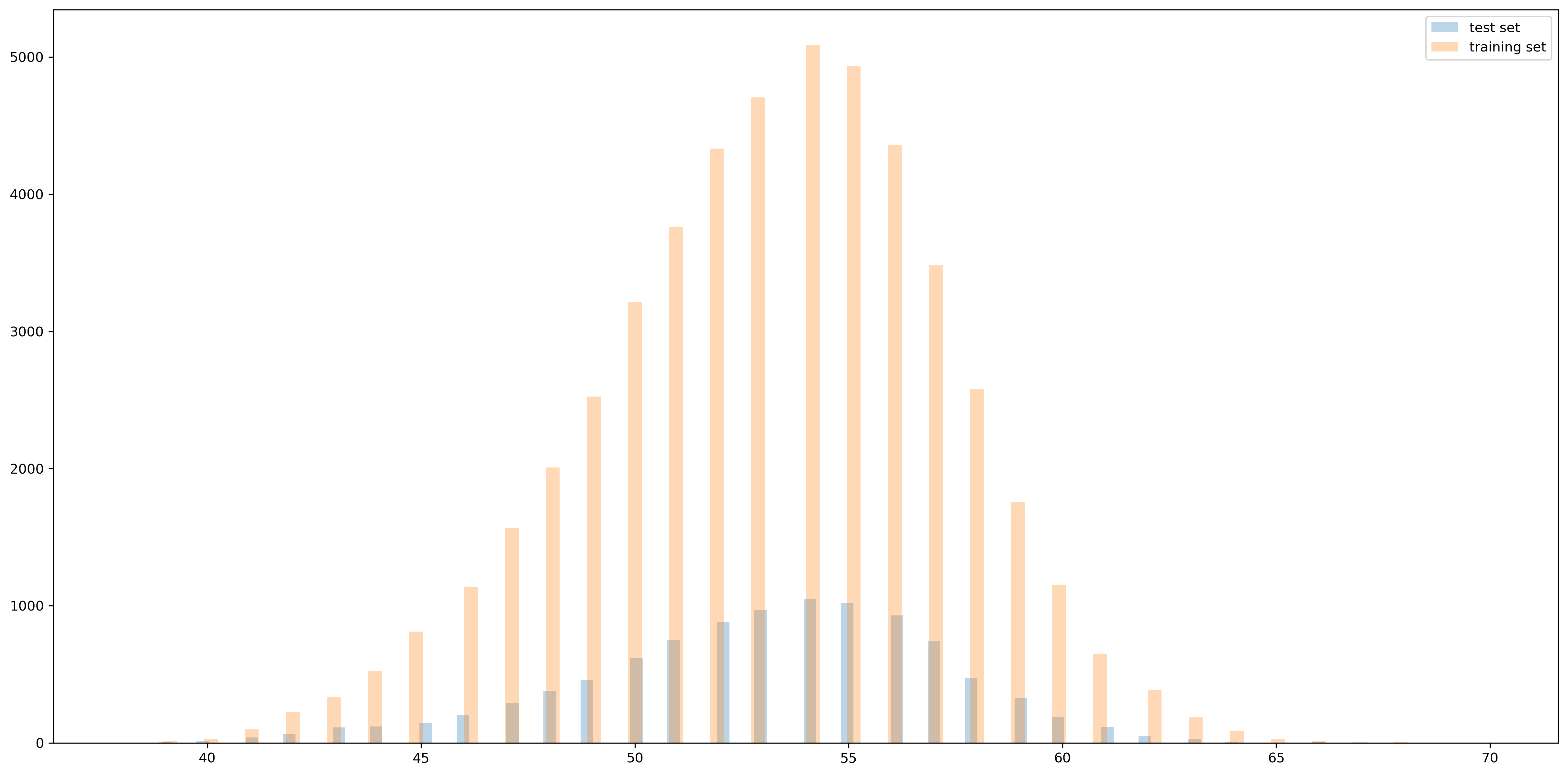}
  \caption{Layer-4}
  \label{fig:sfig54}
\end{subfigure}

\begin{subfigure}{.4\textwidth}
  \centering
  \includegraphics[width=1\linewidth]{cifar_act}
  \caption{All neurons}
  \label{fig:sfig55}
\end{subfigure}

\caption{Histogram of active neurons (nodes in the network) after 50 epochs on the training and on the test set of CIFAR-10 dataset in a 4-layer feed-forward ReLU network with $100$ neurons per layer.}
\label{fig:normal}
\end{figure}

In this section we prove Theorem 3.2. Based on empirical counting (see Fig.~\ref{fig:normal}) we may assume that the number of active nodes follows a normal distribution. Worth mentioning that this assumption is not necessary accurate for active nodes per layer. In a further study we plan to investigate this phenomenon. 
 
\begin{customthm}{3.2}\label{theo2_app}
For $x \sim D$ and a biasless feed-forward ReLU network with input dimension $d_{in}$, $N_{\theta}$ trainable parameters, $w_{max_i}=\max_{w \in \theta_i} |w|$, with the number of active nodes $T(x)$ following a normal distribution $\mathcal{N}(\mu,\sigma)$, the Forbenius norm of \textit{tangent sensitivity} is upper bound by 
\begin{equation}
\nonumber
N_{\theta}  d_{in} \sigma^{2(k-1)}\frac{2^{k-1}}{k^{2k}}\frac{(\Gamma(k/2))^2}{\Pi}(\Psi(-(k-1)/2,1/2,-\mu^2/(2\sigma^2))^2 (\Pi_{i=1}^k w_{max_i})^2.
\end{equation}
where $\Psi$ is Krummer's confluent hypergeometric function.  
\end{customthm}
\begin{proof}
Every positive path should include at least one active node per layer therefore the maximal number of active nodes is $T-k+1$ thus, based on the inequality of the arithmetic and geometric means, the maximal number of positive paths will be $\Pi_{i=1}^k n_i(x)\leq \left(\frac{\sum_{i=1}^k n_i(x)}{k}\right)^k = \left(\frac{T(x)}{k}\right)^k$. As we are interested in the expected number of positive paths, we need $\frac{1}{k^k}\mathbf{E}_{x \sim D}[T(x)^k]$, the $k$-th moment of $T(x)$. Although the computation of higher order moments can be numerically challenging, we only have to derive the absolute moment given that $T(x)\geq 0,\forall x$ thus $\mathbf{E}_{x \sim D}[|T(x)|^k] = \sigma^{k}2^{k/2}\frac{\Gamma((k+1)/2)}{\sqrt{\Pi}}\Psi(-k/2,1/2,-\mu^2/(2\sigma^2)) $\cite{winkelbauer2012moments} where we denote Krummer's confluent hypergeometric function with $\Psi(-k/2,1/2,-\mu^2/(2\sigma^2)) = \sum_{n=0}^{\infty} \frac{(-k/2)^{(n)}}{(1/2)^{(n)} }\frac{ (-\mu/2\sigma^2)^n}{n!}$ and rising factorial with $a^{(n)}=\Pi_{i=1}^n (a-k+1)$. With monotonicity of $T(x)$ and substitutions we get the result. 
\end{proof}

\section{Appendix C}
\label{sec:app_c}

In this section we prove Lemma 3.3 for activation regions and \textit{tangent sample sensitivity} in ReLU networks. 

\begin{customlem}{3.3}\label{lemma1_app}
For each element in an activation region $R(A;\theta)$ \textit{tangent sample sensitivity} is identical. 
\end{customlem}
\begin{proof}
By definition for any element in $R(A;\theta)$ the active neurons per layer are equal and therefore the active paths for any input-output pair is identical thus \textit{tangent sample sensitivity} is the same. 
\end{proof}


Practical calculation of relative frequency of activation regions could be difficult as the number of activation regions are much higher than the size of the sample set. Following the assumption that for the $l$-th hidden unit $\log\frac{p(A_{l}=1|x;\theta)}{1-p(A_{l}=1|x;\theta)} \approx h_{l}(x;\theta)$ than for an input $x$ and an activation pattern $A$ the approximated (membership) probability is   
\begin{equation}
\label{est_p_app}
p(A=\{a_{l};l\in \{1,2,..,N\}\}|x;\theta) \approx \Pi_{l|a_{l}=1} \sigma(h_{l}(x;\theta)) \Pi_{l | a_{l}=-1} (1-\sigma(h_{l}(x;\theta))
\end{equation}
where $\sigma(z) = 1/(1+exp(-z))$ denotes the sigmoid function. To relate the membership probability to the margin of individual neurons let us investigate a single neuron. The margin for a single neuron is defined as the minimal absolute preactivation for a finite set of input: $\rho_{l}(X) := \min_{x \in X} |h_l(x;\theta)|$. Because of the monotonicity of the sigmoid function we can explain the margin in a probabilistic sense with $\hat{\rho}_{l}(X) := \min_{x \in X} |\sigma(h_l(x;\theta)) - 0.5 |$. The connection between $\rho$ and $\hat{\rho}$ depends on the preactivation. For a neuron and input with positive preactivation $\sigma(h_l(x;\theta))$ is larger than $0.5$ similarly for neurons with negative preactivation $\sigma(h_l(x;\theta))$ is smaller than $0.5$ therefore for points inside a region every element in \ref{est_p_app} are larger than $0.5$. It is worth mentioning that it is possible that a point outside a region has higher membership probability than a point inside the region. In a further study we plan to examine more complex estimations of the membership probability. 





\section{Appendix D}
\label{sec:app_d}

\begin{table}
\centering
\begin{tabular}{| c | c | c | c |}
\hline
Layer & \#nodes & \#parameters & Variants\\ \hline
Input layer & 3072 & 0 & \\ \hline
Hidden layer 1 &  100 & $100 \times 3072 + 100$ & -/BN \\ \hline
Hidden layer 1 &  100 & $100 \times 100 + 100$ & -/BN/DO \\ \hline
Hidden layer 1 &  100 & $100 \times 100 + 100$ & -/BN/DO \\ \hline
Hidden layer 1 &  100 & $100 \times 100 + 100$ & -/BN/DO \\ \hline
Output layer &  10 & $10 \times 100 + 10$ & \\ \hline
\end{tabular}
\caption{Network layout for CIFAR-10 dataset. We denote Batch normalization \cite{ioffe2015batch} with BN and Dropout \cite{srivastava2014dropout} with DO.}
\label{tab:net}
\end{table}

We measured the performance of the suggested loss estimations in Section 5 on the CIFAR-10 dataset \cite{krizhevsky2009learning}. We used the training batches as training set and the sixth batch as test set. We implemented simple fully connected ReLU networks in PyTorch \footnote{\url{https://pytorch.org}}. Table~\ref{tab:net} shows the outline of the networks we used in our experiments. We plan to share the python source code. Parameters were initialized uniformly e.g. for the $i$-th layer with $\mathcal{U}(-\sqrt{\frac{1}{N_{i-1}}}, \sqrt{\frac{1}{N_{i-1}}})$. For all experiments we optimized for cross-entropy loss with Stochastic Gradient Descent (SGD) or Adam \cite{kingma2014adam} with batch size of $64$, learning rate of $\alpha=0.05$ and weight decay with $\beta=0.0005$. We evaluated the performance on the test set after every epoch on the training set with cross-entropy loss and accuracy. To compute \textit{tangent sample sensitivity} we saved the preactivations of the hidden nodes in the network per sample as we may calculate posterior probabilities in (eq. 4) based on the preactivations during inference. In addition, the network parameters are available in the model object thus the complexity of empirical \textit{tangent sensitivity} is linear in the size of the sample set with a significant constant for (eq. 7). We measured the quality of the estimated test cross-entropy loss and test accuracy with Mean Absolute Error (MAE), for details see Table~\ref{tab:mae}. The lowest MAE cross-entropy loss and accuracy were achieved by empirical sensitivity (eq. 7) and layer-wise log-norm sensitivity (eq. 5) respectively. 

\begin{table}
\centering
\begin{tabular}{| c | c | c |}
\hline
Estimation & Cross-entropy& Accuracy\\ \hline
Layer-wise log-norm (p=inf) sens. eq.~(5) & 5.3e-4 & 9.2e-3  \\ \hline
Maximal sens. eq.~(6) & 7.7e-4 & 9.8e-3 \\ \hline
Empirical sens. eq.~(7) & 4.2e-4 & 1.1e-2\\ \hline
\end{tabular}
\caption{Mean Absolute Error (MAE) of estimated test cross-entropy loss and test accuracy of ReLU networks on different states on the CIFAR-10 dataset based on layer-wise log-norm ($l_{\infty}) $ (eq. 5), maximal (eq. 6) and empirical \textit{tangent sensitivity} (eq. 7)}
\label{tab:mae}
\end{table}

\end{document}